\documentclass{article}
\usepackage{spconf,amsmath,graphicx}
\usepackage{booktabs}
\usepackage{multirow,balance,cite,color}

\def\x{{\mathbf x}}
\def\L{{\cal L}}
\newcommand{\jie}[1]{#1}

\usepackage{color,soul}
\sethlcolor{cyan}

\let\OLDthebibliography\thebibliography
\renewcommand\thebibliography[1]{
  \OLDthebibliography{#1}
  \setlength{\parskip}{0pt}
  \setlength{\itemsep}{0pt plus 0.3ex}
}

\begin{document}\sloppy

\def\x{{\mathbf x}}
\def\L{{\cal L}}

\title{Semi-Supervised Federated Learning for Keyword Spotting}
%
\name{Enmao Diao, Eric W. Tramel, Jie Ding, Tao Zhang}
\address{
    Amazon, Alexa AI
  }

\maketitle

\begin{abstract}

Keyword Spotting (KWS) is a critical aspect of audio-based applications on mobile devices and virtual assistants. Recent developments in Federated Learning (FL) have significantly expanded the ability to train machine learning models by utilizing the computational and private data resources of numerous distributed devices. However, existing FL methods typically require that devices possess accurate ground-truth labels, which can be both expensive and impractical when dealing with local audio data. In this study, we first demonstrate the effectiveness of Semi-Supervised Federated Learning (SSL) and FL for KWS. We then extend our investigation to Semi-Supervised Federated Learning (SSFL) for KWS, where devices possess completely unlabeled data, while the server has access to a small amount of labeled data. We perform numerical analyses using state-of-the-art SSL, FL, and SSFL techniques to demonstrate that the performance of KWS models can be significantly improved by leveraging the abundant unlabeled heterogeneous data available on devices.

\end{abstract}
\vspace{-0.2cm}
\section{Introduction}
\vspace{-0.2cm}


Keyword spotting (KWS) focuses on identifying pre-specified keywords in audio signals derived from human speech. Recognizing speech commands is crucial for audio-based interactions on mobile devices and virtual assistants \cite{lopez2021deep}. This work investigates the potential of Federated Learning (FL) \cite{mcmahan2017communication,diao2021heterofl} for training a supervised global model on distributed edge devices. In several application domains, including KWS, it may not be feasible for clients to provide precise annotations for unlabeled on-device data. These observations have prompted us to develop an FL framework that utilizes the unlabeled audio streams on distributed clients' devices.

\begin{figure}[tb]
\centering
 \includegraphics[width=0.6\linewidth]{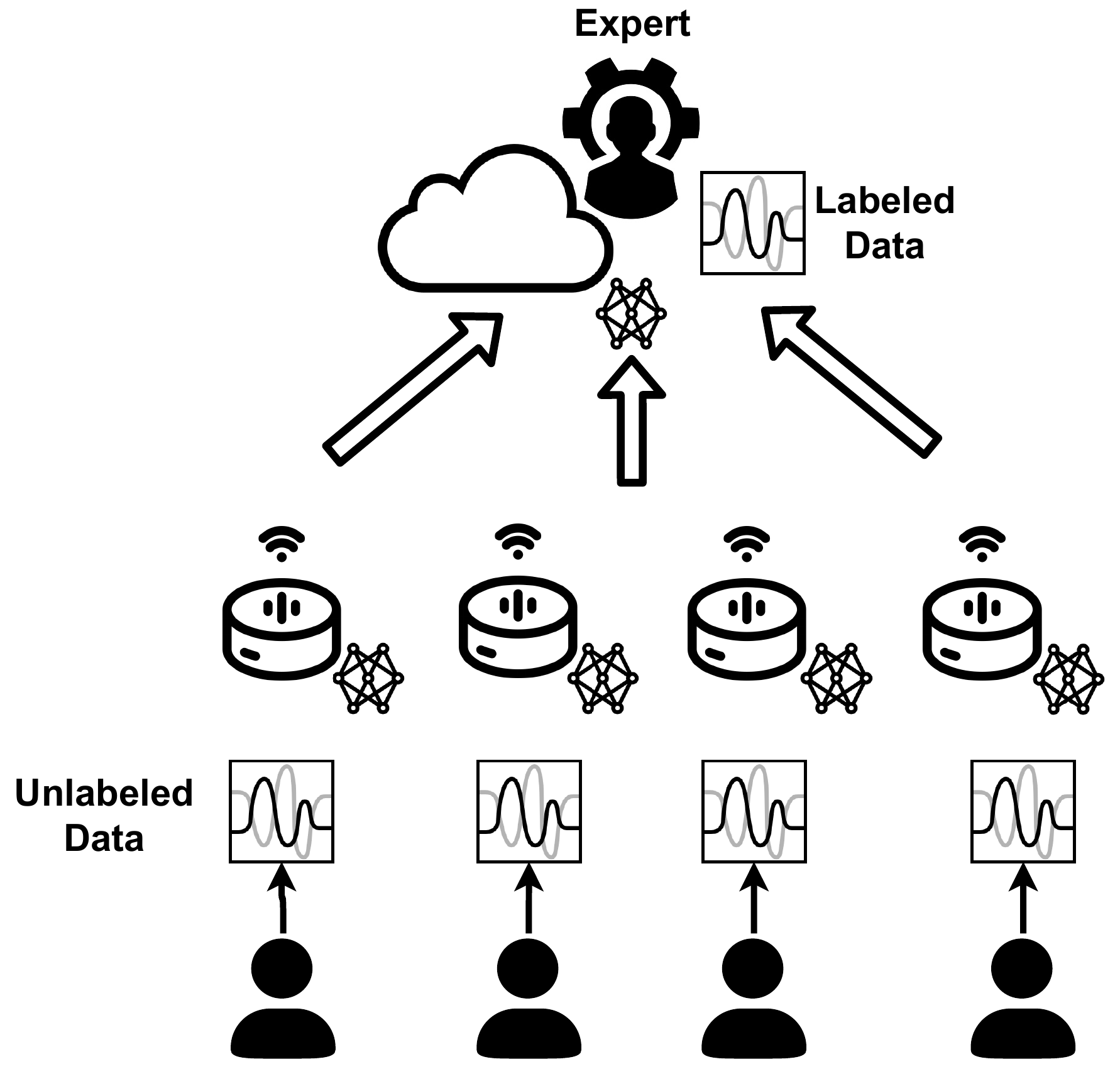}
  \vspace{-0.2cm}
 \caption{An illustration of Semi-Supervised Federated Learning (SSFL) for Keyword Spotting (KWS).}
 \vspace{-0.4cm}
 \label{fig:ssfl}
\end{figure}

In this work, we demonstrate the efficacy of Semi-Supervised Learning (SSL) and Federated Learning (FL) for Keyword Spotting (KWS) and propose a novel SSL and FL integration for Semi-Supervised Federated Learning (SSFL) in KWS. Additionally, we show that the alternate training method used in SSFL can mitigate the Non-IID problem in FL and transfer pre-trained KWS models with unlabeled on-device data. Our research focuses on label-skewed data heterogeneity, where devices may have different keyword distributions. We conduct extensive experiments and ablation studies to assess the effectiveness of data augmentation techniques for KWS to address the SSL and SSFL challenges. Figure \ref{fig:ssfl} demonstrates that leveraging unlabeled on-device data in a distributed manner can significantly enhance the performance of KWS models trained with labeled data.

\begin{figure*}[htb]
\centering
 \includegraphics[width=0.7\linewidth]{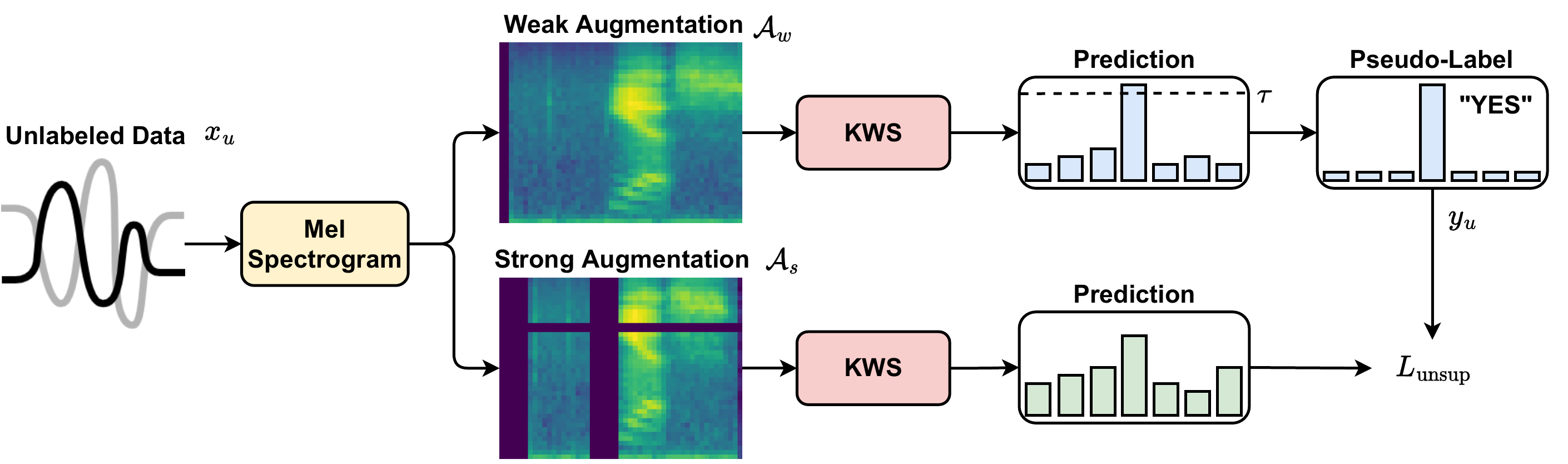}
  \vspace{-0.4cm}
 \caption{An illustration of leveraging unlabeled audio streams in Semi-Supervised Learning (SSL).}
 \vspace{-0.4cm}
 \label{fig:semi}
\end{figure*}

\vspace{-0.3cm}
\section{Related Works}
\vspace{-0.2cm}

Keyword spotting aims to detect predetermined specific words from audio streams \cite{lopez2021deep}. Usually, audio streams are processed locally on users' devices to save computation and communication expenses. Recently, data augmentation and deep neural networks have proved to be effective for KWS \cite{park2019specaugment, rybakov2020streaming}. Semi-Supervised Learning (SSL) refers to the process of training a model using partially labeled data, especially when the amount of labeled data is significantly less than that of unlabeled data. Recently, SSL methods based on image augmentation techniques have been developed for computer vision tasks to achieve state-of-the-art performance \cite{sohn2020fixmatch,diao2021semifl}. Federated Learning (FL) aims to expedite and scale the distributed training of models \cite{mcmahan2017communication}. Several studies have been proposed to reduce the computation and communication costs in FL \cite{mcmahan2017communication,diao2021heterofl}. It has been demonstrated that for speech recognition, FL is advantageous for KWS to exploit local data and computation resources \cite{leroy2019federated}. The integration of Semi-Supervised Learning and Federated Learning is commonly referred to as Semi-Supervised Federated Learning (SSFL) \cite{diao2021semifl}.

\vspace{-0.3cm}
\section{Method}
\vspace{-0.2cm}

\noindent  \textbf{Semi-Supervised Federated Learning} \, In this research, we investigate Semi-Supervised Federated Learning (SSFL) in a scenario where the data available on the device-side is entirely unlabeled, and only the server possesses some labeled data. Previous studies \cite{diao2021semifl} have shown that if the model parameters trained by the server and devices are directly aggregated, the performance may significantly deteriorate when devices train multiple local epochs \cite{mcmahan2017communication}. Therefore, we apply the alternate training technique \cite{diao2021semifl} that employs unlabeled and labeled data iteratively to update models.



We consider a scenario where there are $M$ devices. At the $t$-th round of Federated Learning (FL), $M_t$ active devices upload their trained model $\theta_m^t$ to the server for aggregation, resulting in the aggregated model $\theta^{t}= \frac{1}{M_t} \sum_{m=1}^{M_t}\theta_{m}^t$. The server has access to a small labeled dataset $\mathcal{L} = \{x_{l,i}, y_{l,i}\}^{N_\mathcal{L}}_{i=1}$, which can be utilized to train a server model with the supervised loss $L{\text{sup}}$. Here, $\mathcal{A}_w$ denotes a function of weak data augmentation \cite{sohn2020fixmatch}, $f_{\theta}$ is a parameterized Keyword Spotting (KWS) model, and $\ell$ is the cross-entropy function. The supervised loss is expressed as:
\begin{align}
    L_{\text{sup}} &= \ell(f_{\theta}(\mathcal{A}_w(x_{l})), y_{l}). 
\end{align}
We use an alternate training technique to balance the training of labeled and unlabeled data. In the subsequent iteration, active devices receive the fine-tuned model from the server and generate pseudo-labels for training the unlabeled data $\{x_{u,i}\}^{N_\mathcal{U}}_{i=1}$ using the unsupervised loss $L{\text{unsup}}$. Here, $\mathcal{A}s$ denotes a function of strong data augmentation \cite{sohn2020fixmatch}, and $\tau$ is a threshold value. The unsupervised loss is expressed as:
\begin{align}
    L{\text{unsup}} &= \ell( f_{\theta}(\mathcal{A}s(x_{u})), y_{u}), \\ \nonumber 
    y_{u} &= f_{\theta}(\mathcal{A}_w(x_{u})), 
    \max{\left(y_{u}\right)}\geq\tau.
\end{align}


\noindent \textbf{Data augmentation} \,
Data augmentation is a crucial component of state-of-the-art SSL methods. Prior works on KWS have employed various data augmentation techniques, such as random shifting and resampling of audio streams, addition of background noises, and masking of frequency and time components \cite{park2019specaugment, rybakov2020streaming}. To better leverage SSL methods, we treat mel-spectrograms as images and employ image-based data augmentation techniques such as RandAugment \cite{cubuk2020randaugment}. Mixup data augmentation \cite{berthelot2019mixmatch, diao2021semifl}, which uses a random convex combination to mix labeled and unlabeled data, is another type of SSL method. We conduct extensive ablation studies and demonstrate that appropriate utilization of data augmentation strategies can leverage the abundance of unlabeled data to enhance the performance of KWS models.


\noindent  \textbf{Transfer from pretrained models} \, Classical SSL and SSFL methods assume a small labeled dataset and a large unlabeled dataset, i.e., $N_\mathcal{L} \ll N_\mathcal{U}$. When a large labeled dataset is available, SSFL can adapt pre-trained KWS models to new data domains. We propose fine-tuning pre-trained models $\tilde{\theta}$ with a small labeled dataset on the server and a large unlabeled dataset on devices. The transferred KWS model parameters can be obtained as $\hat{\theta} = \underset{\tilde{\theta}}{\operatorname{argmin}}  , \ell(f_{\tilde{\theta}}(\mathcal{A}_w(x_{l})), y_{l}) + \ell( f_{\tilde{\theta}}(\mathcal{A}_s(x_{u})), y_{u})$, where $\mathcal{A}_w$ and $\mathcal{A}_s$ are functions of weak and strong data augmentation, respectively, and $\ell$ is the cross-entropy loss function.

\vspace{-0.4cm}
\section{Experiments}
\vspace{-0.2cm}
\subsection{Experimental Setup}
\vspace{-0.2cm}

 \noindent  \textbf{Speech Commands datasets} \, We conduct experiments with the public benchmark datasets Speech Commands V1 and V2 datasets \cite{warden2018speech}. The datasets comprise twelve class labels, including ten words (`yes', `no', `up', `down', `left', `right', `on', `off', `stop', and `go'), and two classes (`silence' and `unknown'). The `silence' class is randomly sampled from the background noise, while the `unknown' class comprises the remaining keywords with equal size from the datasets. As our backbone model, we employ Temporal Convolution ResNet18 (TC-ResNet18) \cite{choi2019temporal}. Our experiments involve $100$ devices, and we maintain a constant ratio of active devices per communication round $C = 0.1$ in all experiments \cite{mcmahan2017communication}. We equally allocate data examples for the IID data partition. For a balanced Non-IID data partition, we adopt the method outlined in earlier work \cite{diao2021heterofl, diao2021semifl} and set $K=2$ to simulate label-skewed data heterogeneity. We sample data from a Dirichlet distribution $\operatorname{Dir(\alpha)}$ \cite{diao2021semifl} for an unbalanced Non-IID data partition. We conduct four random experiments with different seeds and report the standard deviation in parentheses for tables and by error bars in figures.


\noindent  \textbf{Wakeword dataset} \, In this study, we conduct an evaluation of SSL and SSFL techniques on an internal audio dataset that has been de-identified. The dataset comprises approximately 14K hours of training data and 1.5K hours of test data, and is used for the purpose of wakeword detection. Wakeword detection is a task that involves using an on-device model to identify a specified keyword and thereby initiate device activation. Specifically, our experiment focuses on a binary classification task with two output classes: wakeword detected and wakeword absent. To this end, we represent audio streams as mel-spectrograms and employ a Convolutional Neural Network (CNN) with multiple convolution and max-pooling layers. To evaluate the performance of the resulting models, we use the Relative False Reject Rate at a False Accept Rate (Relative FRR@FAR) metric. This metric is defined as the ratio of the FRR of the test model to that of the baseline model, given a fixed FAR of the baseline model. Consequently, a Relative FRR@FAR value less than one is preferred. We conduct four random experiments with different seeds, and the standard deviation is within $0.01$. 

\vspace{-0.4cm}
\subsection{Experimental Results}
\vspace{-0.2cm}

\noindent  \textbf{Semi-Supervised Federated Learning} \, In this study, we compare our findings with two baseline methods, namely, 'Fully Supervised' and 'Partially Supervised' in the centralized setting. `Fully Supervised' refers to the approach of training the model with all data being labeled, while 'Partially Supervised' denotes training the model with partially labeled data. To evaluate the results, we use Accuracy in Table \ref{tab:result_1} and Relative FRR@FAR in Table \ref{tab:result_2}. Our analysis reveals that SSL and SSFL techniques outperform the 'Partially Supervised' case and perform comparably to the `Fully Supervised' case. Specifically, with $250$ and $2500$ labeled data from the Speech Command datasets, we achieve approximately $40\%$ and $10\%$ improvement over the `Partially Supervised' case in Accuracy, respectively. With $2500$ labeled wakeword streams, we improve the performance of the pretrained models and the 'Partially Supervised' case by around $0.2$ and $0.1$ in Relative FRR@FAR, respectively. Overall, our results demonstrate the effectiveness of training KWS models from scratch or transferring from pretrained models with abundant unlabeled on-device data to improve performance.

\begin{table}[htb]
\centering
\caption{Comparison of SSL and SSFL with baselines for the public Speech Commands datasets using Accuracy ($\uparrow$).}
\label{tab:result_1}
\resizebox{1\columnwidth}{!}{
\begin{tabular}{@{}ccccccc@{}}
\toprule
\multicolumn{3}{c}{Dataset}                                                               & \multicolumn{2}{c}{Speech Commands V1} & \multicolumn{2}{c}{Speech Commands V2} \\ \midrule
\multicolumn{3}{c}{\jie{Proportion of Supervised Data}}                                                & \jie{$\approx $1\%}               & \jie{$\approx $10\%}           & \jie{$\approx $1\% }              & \jie{$\approx $ 10\%}           \\ \midrule
\multicolumn{3}{c}{Fully Supervised}                                                    & \multicolumn{2}{c}{99.7(0.0)}        & \multicolumn{2}{c}{99.6(0.0)}        \\ \midrule
\multicolumn{3}{c}{Partially Supervised}                                                & 62.4(0.7)         & 90.3(0.2)        & 60.2(0.9)         & 89.7(0.1)        \\ \midrule
\multicolumn{3}{c}{Semi-Supervised}                                                     & 95.7(0.2)         & 97.8(0.0)        & 96.5(0.1)         & 98.1(0.1)        \\ \midrule
\multirow{3}{*}{Non-IID, $K=2$}                     & Parallel                   & FL   & \multicolumn{2}{c}{68.2(1.7)}        & \multicolumn{2}{c}{77.9(2.4)}        \\ \cmidrule(l){2-7} 
                                                    & \multirow{2}{*}{Alternate} & FL   & \multicolumn{2}{c}{98.3(0.1)}        & \multicolumn{2}{c}{97.6(0.0)}        \\ \cmidrule(l){3-7} 
                                                    &                            & SSFL & 88.6(0.3)         & 94.8(0.1)        & 86.9(0.4)         & 93.0(0.1)        \\ \midrule
\multirow{3}{*}{Non-IID, $\operatorname{Dir}(0.1)$} & Parallel                   & FL   & \multicolumn{2}{c}{97.6(0.2)}        & \multicolumn{2}{c}{97.1(0.2)}        \\ \cmidrule(l){2-7} 

                                                    & \multirow{2}{*}{Alternate} & FL   & \multicolumn{2}{c}{98.3(0.2)}        & \multicolumn{2}{c}{98.1(0.0)}        \\ \cmidrule(l){3-7} 
                                                    &                            & SSFL & 90.1(0.0)         & 95.6(0.1)        & 89.1(0.2)         & 94.2(0.2)        \\ \midrule
\multirow{3}{*}{Non-IID, $\operatorname{Dir}(0.3)$} & Parallel                   & FL   & \multicolumn{2}{c}{99.0(0.1)}        & \multicolumn{2}{c}{98.7(0.1)}        \\ \cmidrule(l){2-7} 
                                                    & \multirow{2}{*}{Alternate} & FL   & \multicolumn{2}{c}{99.0(0.1)}        & \multicolumn{2}{c}{98.8(0.0)}        \\ \cmidrule(l){3-7} 
                                                    &                            & SSFL & 93.3(0.4)         & 96.2(0.0)        & 93.1(0.1)         & 95.4(0.1)        \\ \midrule
\multirow{3}{*}{IID}                                & Parallel                   & FL   & \multicolumn{2}{c}{99.6(0.0)}        & \multicolumn{2}{c}{99.3(0.0)}        \\ \cmidrule(l){2-7} 
                                                    & \multirow{2}{*}{Alternate} & FL   & \multicolumn{2}{c}{99.5(0.0)}        & \multicolumn{2}{c}{99.3(0.0)}        \\ \cmidrule(l){3-7} 
                                                    &                            & SSFL & 95.3(0.2)         & 97.8(0.0)        & 95.9(0.1)         & 97.7(0.0)        \\ \bottomrule
\end{tabular}
}
\vspace{-0.3cm}
\end{table}



\begin{figure}[htb]
\centering
 \includegraphics[width=1\linewidth]{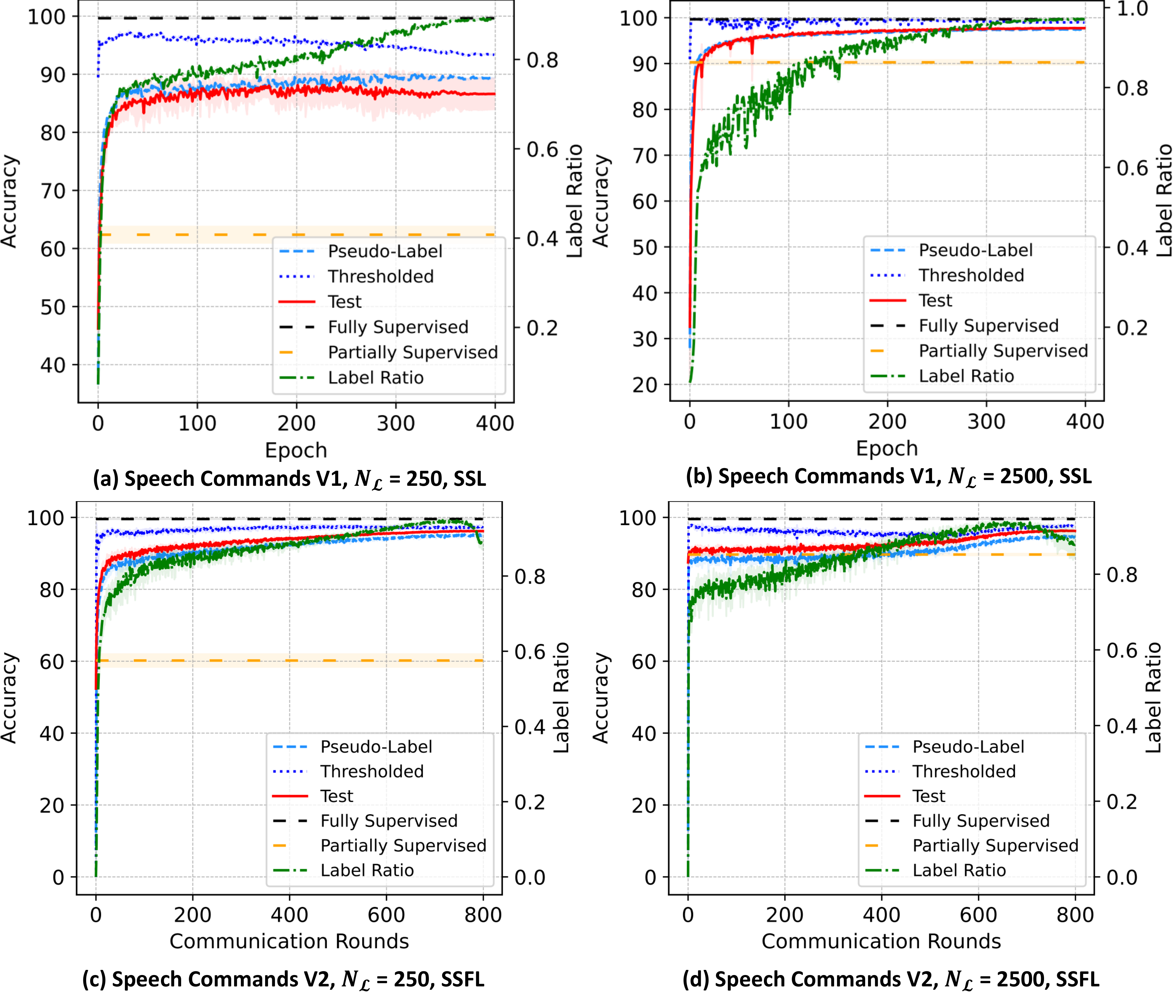}
  \vspace{-0.8cm}
 \caption{Learning curves of SSL and SSFL for TC-ResNet18 with Speech Commands datasets and $N_\mathcal{L}=\{250,2500\}$.}
 \vspace{-0.4cm}
 \label{fig:lc}
\end{figure}

In this study, we present the learning curves of SSL and SSFL as depicted in Figure \ref{fig:lc}. In addition to Test Accuracy, we provide visual representations of the outcomes of `Fully Supervised' and `Partially Supervised' techniques, the accuracy of pseudo-labels before (Label Accuracy) and after (Threshold Accuracy) employing a confidence threshold, and the proportion of unlabeled data above the confidence threshold (Label Ratio). The findings demonstrate an increase in the Accuracy of generated labels subsequent to the filtration of pseudo-labels employing a confidence threshold. Additionally, the labeling ratio increases, indicating a reduction in the prediction's entropy.

\begin{table}[htb]
\centering
\caption{Comparison of SSL and SSFL with baselines for wakeword dataset using Relative FRR@FAR ($\downarrow$) with respect to pretrained models.}
\label{tab:result_2}
\resizebox{0.5\columnwidth}{!}{
\begin{tabular}{@{}cccc@{}}
\toprule
\multicolumn{3}{c}{\jie{Proportion of of Supervised Data}} & \jie{$\approx$ 3\%} \\ \midrule
\multicolumn{3}{c}{Fully Supervised} & 0.80 \\ \midrule
\multicolumn{3}{c}{Partially Supervised} & 0.90 \\ \midrule
\multicolumn{3}{c}{Semi-Supervised} & 0.81 \\ \midrule
\multirow{3}{*}{Imbalanced} & Parallel & FL & 0.85 \\ \cmidrule(l){2-4} 
 & \multirow{2}{*}{Alternate} & FL & 0.81 \\ \cmidrule(l){3-4} 
 &  & SSFL & 0.85 \\ \bottomrule
\end{tabular}
}
\vspace{-0.3cm}
\end{table}

\noindent  \textbf{Alternate training for heterogeneous on-device data } We compare parallel and alternate training for Federated Learning (FL) using Speech Commands V1 and V2 datasets, as presented in Tables \ref{tab:result_1} and \ref{tab:result_2}. `Parallel' training denotes the traditional combination of Semi-Supervised Learning (SSL) and FL, while `Alternate' training involves the server and devices training in an alternating manner \cite{diao2021semifl}. Our results demonstrate that alternate training significantly improves the performance of FL in Non-IID data distributions. Specifically, in the most heterogeneous case (Non-IID, $K=2$), alternate training outperforms parallel training by $30\%$ and $20\%$ in Accuracy for Speech Commands V1 and V2, respectively. Furthermore, when transferring from pretrained models with highly imbalanced on-device data, we observe that alternate training outperforms parallel training by $0.04$ in Relative False Rejection Rate (FRR) at a given False Acceptance Rate (FAR). The findings suggest that a small amount of labeled IID data at the server can effectively address Non-IID data partition in the FL setting.

\begin{figure}[htb]
\centering
 \includegraphics[width=1\linewidth]{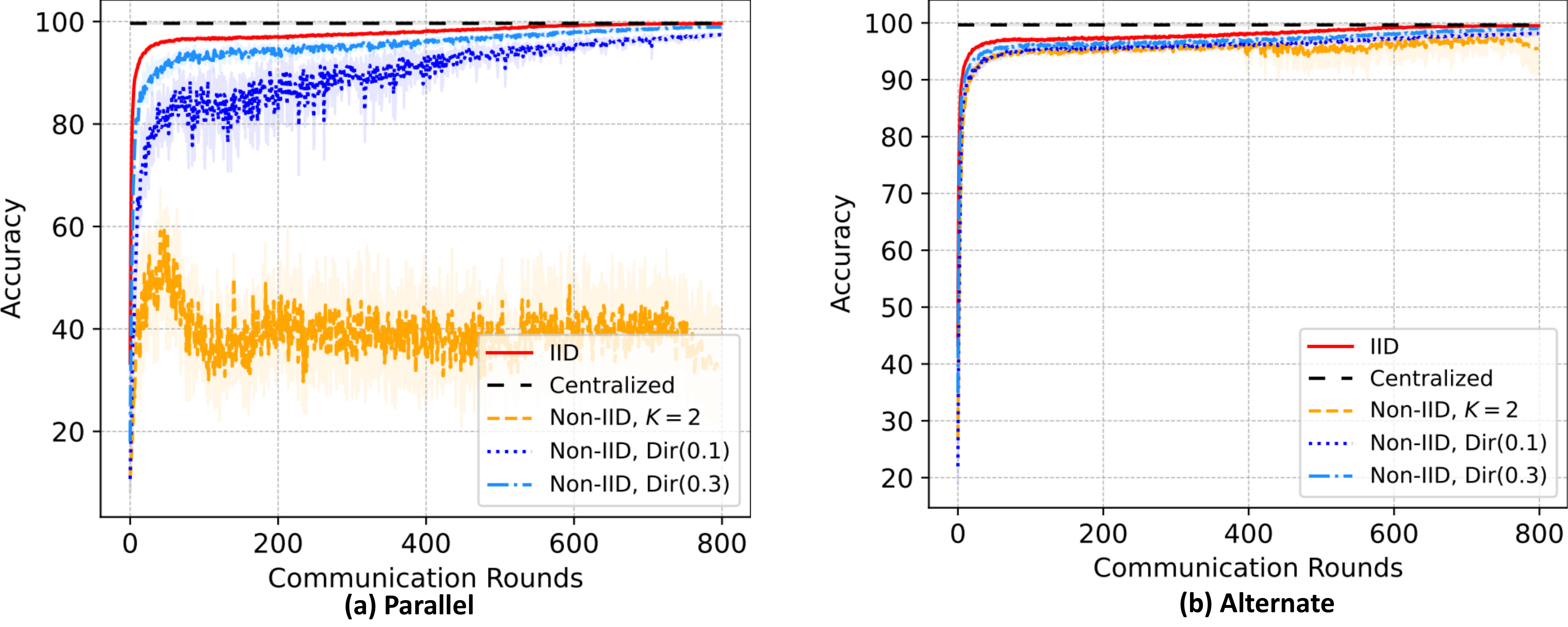}
  \vspace{-0.8cm}
 \caption{Comparison between `Parallel' and `Alternate' training for heterogeneous on-device data.}
 \label{fig:alter}
 \vspace{-0.4cm}
\end{figure}


\noindent \textbf{Data augmentation} \, Our investigation focuses on four data augmentation strategies, namely `BasicAugment,' `SpecAugment,' `RandAugment,' `RandAugment (Selected),' and `MixAugment.' `BasicAugment' involves time shift, resampling, and random background noise, as proposed by Rybakov et al.\cite{rybakov2020streaming}. `SpecAugment' utilizes masking on frequency and time components of mel-spectrograms, as proposed by Park et al.\cite{park2019specaugment}. `RandAugment' employs image-based augmentation methods proposed by Cubuk et al.\cite{cubuk2020randaugment}, while `RandAugment (Selected)' uses selected image-based augmentation strategies. `MixAugment' utilizes a random convex combination to mix labeled and unlabeled data, as proposed by Berthelot et al.\cite{berthelot2019mixmatch}. We conduct ablation studies for strong data augmentation and loss functions, as presented in Table \ref{tab:ablation}. Specifically, we use `BasicAugment' for weak data augmentation $\mathcal{A}_w$ and experiment with various combinations of strong data augmentation methods $\mathcal{A}_s$. Our findings demonstrate that using strong data augmentation can significantly enhance the performance of SSL methods. Moreover, we observe that image-based augmentation methods do not outperform `SpecAugment,' and `MixAugment' can be combined with `SpecAugment' to further improve performance.

\vspace{-0.4cm}

\begin{table}[htbp]
\centering
\caption{Ablation studies of strong data augmentation methods used in SSL for KWS models with $N_{\mathcal{L}} = 250$.}
\label{tab:ablation}
\resizebox{1\columnwidth}{!}{
\begin{tabular}{@{}ccc@{}}
\toprule
Strong Augmentation $\mathcal{A}_s$   & Speech Commands V1 & Speech Commands V2 \\ \midrule
BasicAug                              & 86.1(0.6)          & 85.2(0.6)          \\
BasicAug, RandAug                     & 90.4(0.2)          & 89.1(0.2)          \\
BasicAug, RandAug (Selected)          & 89.8(0.8)          & 89.9(0.3)          \\
BasicAug, SpecAug                     & 95.2(0.3)          & 95.8(0.2)          \\
BasicAug, SpecAug, RandAug (Selected) & 94.4(0.3)          & 94.0(0.1)          \\
BasicAug, SpecAug, MixAug             & \textbf{95.7(0.2)} & \textbf{96.5(0.1)} \\ \bottomrule
\end{tabular}
}
\end{table}

\vspace{-0.5cm}
\section{Conclusion}
\vspace{-0.3cm}
This work studie the efficacy of SSFL methods for KWS and different strong data augmentation techniques. Our experiments suggest that a small amount of labeled data on the server can enable training from scratch or transfer from pretrained models, leveraging heterogeneous unlabeled on-device data.

\vspace{-0.6cm}
\balance
\bibliographystyle{IEEEbib}
\bibliography{References}

\end{document}